\documentclass[10pt,twocolumn,letterpaper]{article}

\usepackage{graphicx}
\usepackage{makecell}
\usepackage{comment}
\usepackage{pifont}

\usepackage[pagenumbers]{cvpr} 

\definecolor{cvprblue}{rgb}{0.21,0.49,0.74}
\definecolor{wh}{rgb}{1,0.,0.}
\usepackage[pagebackref,breaklinks,colorlinks,allcolors=cvprblue]{hyperref}

\def\paperID{***} 
\def\confName{CVPR}
\def\confYear{2026}

\title{MoniRefer: A Real-world Large-scale Multi-modal Dataset based on Roadside Infrastructure for 3D Visual Grounding}

\begin{document}
\renewcommand{\thefootnote}{\fnsymbol{footnote}}
\author {Panquan Yang$^1$\hspace{4mm}
Junfei Huang$^1$\hspace{4mm}
Zongzhangbao Yin$^1$\hspace{4mm}
Yingsong Hu$^1$\hspace{4mm}\\
Anni Xu$^1$\hspace{4mm}
Xinyi Luo$^1$\hspace{4mm}
Xueqi Sun$^1$\hspace{4mm}
Hai Wu$^2$\hspace{4mm}
Sheng Ao$^{1}$\footnotemark[2]\hspace{4mm}\\
Zhaoxing Zhu$^1$\hspace{4mm}
Chenglu Wen$^1$\hspace{4mm}
Cheng Wang$^{1}$\footnotemark[2]\hspace{4mm}\\
$^1$Xiamen University
\hspace{4mm}
$^2$Pengcheng Laboratory\\
}

\maketitle
\footnotetext[2]{Corresponding author.}
\begin{figure*}[ht]
\centering
\includegraphics[width=1.0\textwidth]{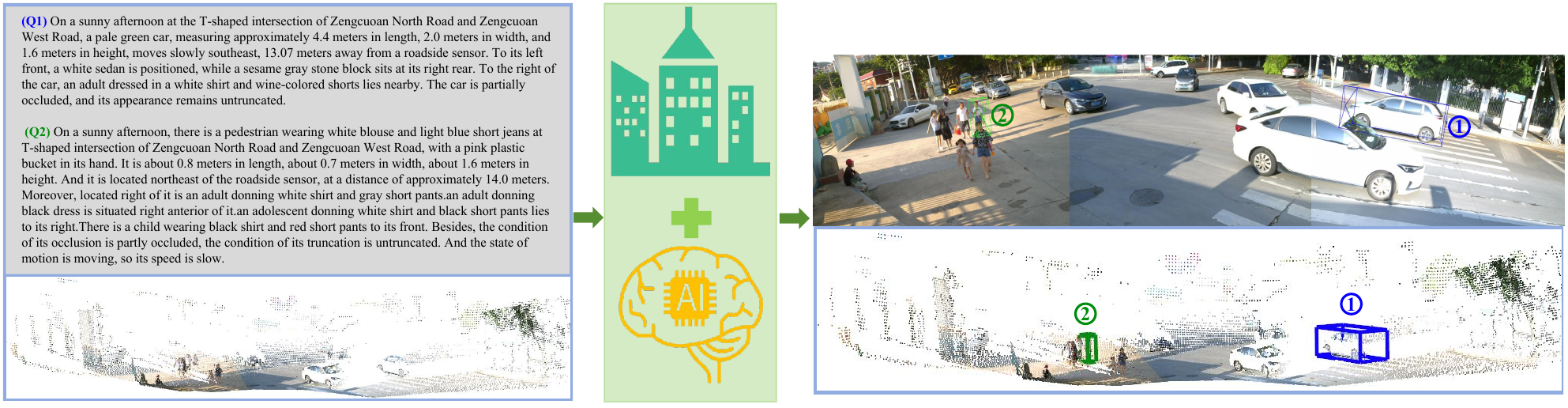}
\caption{\textbf{Introduction to 3D Visual Grounding for Outdoor Monitoring Scenarios.} Moni3DVG integrates multi-modal visual data collected from roadside infrastructure sensors with natural language descriptions to localize the referred object in highly dynamic and complex traffic environments. For clearer visualization, the localization results are also presented in the corresponding images.}
\label{fig:3dvg_task}
\vspace{-4mm}
\end{figure*}
\begin{abstract}
3D visual grounding aims to localize the object in 3D point cloud scenes that semantically corresponds to given natural language sentences. It is very critical for roadside infrastructure system to interpret natural languages and localize relevant target objects in complex traffic environments. However, most existing datasets and approaches for 3D visual grounding focus on the indoor and outdoor driving scenes, outdoor monitoring scenarios remain unexplored due to scarcity of paired point cloud-text data captured by roadside infrastructure sensors. In this paper, we introduce a novel task of \textbf{3D Visual Grounding for Outdoor Monitoring Scenarios}, which enables infrastructure-level understanding of traffic scenes beyond the ego-vehicle perspective. To support this task, we construct \textbf{MoniRefer}, the first real-world large-scale multi-modal dataset for roadside-level 3D visual grounding. The dataset consists of about 136,018 objects with 411,128 natural language expressions collected from multiple complex traffic intersections in the real-world environments. To ensure the quality and accuracy of the dataset, we manually verified all linguistic descriptions and 3D labels for objects. Additionally, we also propose a new end-to-end method, named \textbf{Moni3DVG}, which utilizes the rich appearance information provided by images and geometry and optical information from point cloud for multi-modal feature learning and 3D object localization. Extensive experiments and ablation studies on the proposed benchmarks demonstrate the superiority and effectiveness of our method. Our dataset and code will be released.

\end{abstract}    
\section{Introduction}
3D visual grounding(3DVG) aims at localizing a referred target object in 3D scenes based on a given referring expressions. By establishing an intrinsic connection between linguistic expressions and corresponding visual elements, it enables intelligent systems and embodied agents to possess human-like comprehensive understanding and reasoning ability.3DVG is emerging as a frontier research direction across various domains, including autonomous driving~\cite{pointrcnn,pointpillars,virconv,voxelnext}, embodied intelligence~\cite{robomind,ario,x-embodiment}, vision-language-action~\cite{improvingvla,openvla,momanipvla}, vision-and-language navigation~\cite{vln,duet,aerialvln}, and surveillance~\cite{dair-v2x,infradet3d,rope3d,rcooper}.

In recent years, visual grounding in 2D scenes~\cite{a-attn, mdetr, dynamic-mdetr,fast_accurate,dga,mattnet,cm-att-e} has made significant progress, due to the availability of a large amount of datasets~\cite{refcoco+, flickr30k, refcocog}. However, these methods and datasets are constrained to the 2D domain and cannot capture the true 3D content of objects, making it difficult to directly extend these 2D approaches to the 3D setting. Recent many efforts~\cite{instancerefer,3d-sps,3dvg-transformer,eda,butd_detr,tsp3d} have primarily focused on indoor benchmarks~\cite{scanrefer,referit3d}, where sensing range of RGB-D camera is up to 5 meters and the pre-scanned indoor scenarios is static. It has certain limitations in handling dynamic, complex, large-scale and ever-changing outdoor scenarios. 

Therefore, several works~\cite{talk2car,mssg,mono3dvg,nugrounding} have investigated 3DVG for autonomous driving by leveraging paired linguistic descriptions and visual data, enabling automated vehicles to identify referred targets in driving scenes. Other efforts~\cite{wildrefer} have explored human-centric environments, revealing significant potential for the advancement of human–robot interaction. However, distinct from these settings, 3D visual grounding from roadside surveillance perspectives remains unexplored and faces several key challenges: (1) Data scarcity. LiDAR–text paired data is extremely limited in roadside monitoring scenarios. Most outdoor 3DVG datasets are derived from autonomous driving benchmarks~\cite{nuscenes,kitti}, which contain relatively few objects and focus primarily on vehicles, while underrepresenting vulnerable road users (VRUs) such as pedestrians and cyclists that are critical for situational awareness. (2) Limited linguistic diversity and richness: Existing descriptions are typically brief and coarse, lacking detailed references to object attributes and contextual relationships. Richer and more discriminative language is crucial for resolving object ambiguities in crowded or complex scenes. (3) Perspective inconsistency: the roadside sensors is usually higher and perspectives are typically downward-looking, leading to inconsistent data distribution. In vehicle-mounted systems, limited receptive fields and blind spots caused by occlusion degrade localization accuracy, e.g., ghost probes at busy traffic intersections. In contrast, roadside sensors provide a broader and more stable field of view that enhances perception reliability and safety in intelligent transportation systems. Consequently, existing 3DVG datasets and methods often struggle to generalize effectively to outdoor monitoring environments.

To tackle these challenges, we introduce a novel task of \textbf{3D Visual Grounding for Outdoor Monitoring Scenarios}, which utilize linguistic descriptions and visual data collected from roadside infrastructure sensors to localize target objects in 3D space, as depicted in Fig.\ref{fig:3dvg_task}. To support this task, We build the first real-world large-scale multi-modal dataset for roadside-level 3D visual grounding, termed \textbf{MoniRefer}, which contains 411,128 natural language descriptions of 136,018 objects collected from multiple complex traffic intersections in the real world. Each object is annotated with over 10 rich semantic attributes to enhance linguistic diversity and contextual understanding. To construct a high-quality dataset, both instance bounding boxes and linguistic descriptions generated through human annotation and ChatGPT-assisted, are refined and manually verified. MoniRefer fills the gap in 3DVG research under real-world roadside surveillance settings and provides a strong data foundation for future studies on intelligent transportation and smart city cognitive intelligence.

Furthermore, we propose \textbf{Moni3DVG}, an end-to-end framework that utilizes appearance, optical and geometric features from visual data and linguistics expression to identify the referred object in roadside-view point clouds. The core idea of our method is to identify the 3D bounding box that best matches the natural language description by selecting the candidate point closest to the center of the referred object. Comprehensive experiments on MoniRefer demonstrate the effectiveness and superiority of our approach, establishing a new benchmark for 3D visual grounding in outdoor monitoring scenarios.

In summary, our main contributions are as follow:
\begin{itemize}[leftmargin=2em]
\item We introduce a novel task of 3D visual grounding task for roadside monitoring scenarios that leverages diverse natural language expressions and multi-modal visual data collected from roadside infrastructure sensors, where the LiDAR and cameras are stationary relative to their surroundings.

\item We build the first large-scale multi-modal dataset, termed MoniRefer, for roadside-level 3D visual grounding, featuring 136,018 objects collected from multiple complex traffic intersections, along with 411,128 rich and diverse natural language descriptions, contributing a strong data foundation for advancing future studies in intelligent transportation and smart city cognitive intelligence.

\item We propose an end-to-end method, dubbed as Moni3DVG, which efficiently aggregates appearance, optical, and geometric information for multi-modal learning and 3D object localization, achieving high-performance roadside-level 3D visual grounding. 

\item We provide a sufficient benchmark. Extensive experimental results demonstrate that our proposed end-to-end method achieves state-of-the-art performance on the MoniRefer dataset.
\end{itemize}

\begin{table*}[ht]
\centering
\caption{\textbf{Comparison of 3D visual grounding datasets, } where "T.S." indicates time series, "Avg." means average, "Fur." denotes furniture, "Photog." refers to photogrammetry, "LLM" means large language model, "MLLM" represents multi-modal LLM, "H/V" refers to human/vehicle, "VRU/V/O" is interpreted as vulnerable road user/vehicle/obstacle ,"Hum.-cent" corresponds to human-centric, "Veh.-cent" indicates vehicle-centric, "Infra.-Level" means infrastructure-level.}
\label{tab:1}
  \setlength{\tabcolsep}{2pt}
    \resizebox{1.\linewidth}{!}{
\begin{tabular}{lcccccccccccccc}
\Xhline{2\arrayrulewidth}
Dataset & Sensor & Language Form & Range & Target & Outdoor & T.S. & Objects & Expressions & Avg.Length & Vocab & Views & Scene Nums & Task Level & Label \\ 
\Xhline{2\arrayrulewidth}
ScanRefer~\cite{scanrefer}   & RGB-D     & Manual       & 10m  & Fur.        & \ding{55} & \ding{55} & 11.0k    & 51.6k    & 20.27 & 4,197  & - & 800   & Room-Level   & 3D BBox \\
Nr3D~\cite{referit3d}        & RGB-D     & Manual       & 10m  & Fur.        & \ding{55} & \ding{55} & 5.9k   & 41.5k  & 11.4 & 6,951 & - & 641   & Room-Level   & 3D BBox \\
Sr3D~\cite{referit3d}        & RGB-D     & Templated      & 10m  & Fur.        & \ding{55} & \ding{55} & 8.9k   & 83.5k  & -    & 196   & - & 1,273 & Room-Level   & 3D BBox \\
SUNRefer~\cite{sunrefer}     & RGB-D     & Manual       & 10m  & Fur.        & \ding{55} & \ding{55} & 7.7k   & 38.5k  & 16.3 & 5,279  & - & 7,699  & Room-Level   & 3D BBox \\
CityRefer~\cite{cityrefer}   & UAV Photog.   & Manual       & -    & Variety       & \ding{51} & \ding{55} & 5.9k   & 35.2k  & -    & 6,683 & - & -     & City-Level   & 3D BBox \\
STRefer~\cite{wildrefer}     & LiDAR+RGB       & Manual       & 30m  & Human        & \ding{51} & \ding{51} & 3.6k   & 5.5k   & -    & -     & 1 & 662   & Hum.-Cent   & 3D BBox \\
LifeRefer~\cite{wildrefer}   & LiDAR+RGB       & Manual       & 30m  & Human
& \ding{51} & \ding{51} & 11.8k  & 25.4k  & -    & -     & 1 & 3,172 & Hum.-Cent   & 3D BBox \\
Mono3DRefer~\cite{mono3dvg}  & RGB       & LLM+Manual  & 102m & H/V         & \ding{51} & \ding{55} & 8.2k   & 41.1k  & 53.24 & 5,271 & 1 & 2,025 & Veh.-Cent   & 2D/3D BBox \\
Talk2Car~\cite{talk2car}     & RGB       & Manual       & -    & H/V        & \ding{51} & \ding{55} & 10.5k  & 11.9k  & 11.01 & -     & 1 & 9,217  & Veh.-Cent   & 2D BBox \\
Talk2Car-3D~\cite{talk2lidar}     & LiDAR+RGB       & Manual       & -    & H/V        & \ding{51} & \ding{55} & -  & 8.4k  & - & -     & 1 & 5,534  & Veh.-Cent   & 3D BBox \\
Talk2Lidar~\cite{talk2lidar} & LiDAR+RGB       & MLLM+LLM       & -    & H/V         & \ding{51} & \ding{55} & -      & 59.2k  & -    & -     & 6 & 6,419  & Veh.-Cent   & 3D BBox \\
NuPrompt~\cite{nuprompt}     & RGB       & Manual+LLM   & -    & H/V         & \ding{51} & \ding{51} & -      & 40.1k  & -    & -     & 6 & 34,149 & Veh.-Cent   & 3D BBox \\
\textbf{Ours}                & \textbf{LiDAR+RGB} & \textbf{Manual+LLM} & \textbf{50m} & \textbf{VRU/V/O} & \ding{51} & \ding{51} & \textbf{136.0k} & \textbf{411.1k} & \textbf{126.8} & \textbf{8,478} & \textbf{3} & \textbf{12,085} & \textbf{Infra.-Level} & \textbf{2D/3D BBox} \\
\Xhline{2\arrayrulewidth}
\end{tabular}}
\end{table*}

\section{Related Work}

\noindent\textbf{2D Visual Grounding.}
The goal of 2D visual grounding(2DVG) is to localize a specific region within an image according to a natural language expression. Over the past decade, numerous datasets~\citep{refcoco+, flickr30k, refcocog, referitgame} and methods~\citep{a-attn, mdetr, dynamic-mdetr, cmcg, dq-detr} have driven significant progress in this area. Existing approaches can be broadly categorized into two paradigms: two-stage and one-stage methods. Two-stage methods~\citep{mc, vc, cmn, rvgtree,dga,mattnet,cm-att-e} first generate a set of region proposals and then select the proposal with the highest region-text matching score as the final prediction. To avoid proposal generation and reduce computational cost, one-stage methods~\citep{fast_accurate,resc,look-byl,real_time,mmca} have been developed, which densely fuse image and language features to directly regress the target bounding box. However, directly extending these methods to 3D scenes is not feasible due to the sparse, unordered, and irregular nature of point clouds~\citep{pointnet, pointnet++}, as well as the lack of accurate 3D geometric information in 2D images, which limit their applicability in real-world scenarios such as autonomous driving and roadside monitoring.

\noindent\textbf{3D Visual Grounding.}
The objective of 3D visual grounding (3DVG) is to localize a target object in a 3D scene based on a linguistic description. Early works such as ScanRefer~\cite{scanrefer} and ReferIt3D~\cite{referit3d}, derived from the ScanNet~\cite{scannet}, have greatly promoted the progress of 3DVG in indoor environments. Similar to 2DVG, 3DVG methods are also divided into two-stage and single-stage approaches. Two-stage methods follow a detection-then-matching paradigm. Scanrefer~\cite{scannet} firstly utilize a pretrained 3D object detector~\cite{votenet} to generate a set of object proposals and then match these proposals with given linguistic description to identify target object. Instancerefer~\cite{instancerefer} reformulates this task as instance matching problem. 3DJCG~\cite{3djcg} and UniT3D~\cite{unit3d} introduce dense captioning as an auxiliary task. Transformer-based models ~\cite{transrefer3d,3dvg-transformer,mvt} have also been explored. In contrast, single-stage method 3D-SPS ~\cite{3d-sps} directly deduces target object from point clouds, without relying on the performance of 3D object detector. BUTD-DETR~\cite{butd_detr} encodes bounding box proposal and decodes objects from contextual features.EDA~\cite{eda} introduces a text decoupling module to parse linguistic sentence into multiple semantic components.MCLN~\cite{mcln} utilizes a 3D referring expression segmentation task to facilitate collaborative Learning.TSP3D~\cite{tsp3d} leverages multi-level sparse convolutional architecture to achieve efficient 3DVG. These approaches and datasets are all based on RBG-D scans in static indoor scenarios, where the perception range of camera is up to 5 meters. Therefore, Cityrefer~\cite{cityrefer} proposes to perform city-level 3DVG.Wildrefer~\cite{wildrefer} focuses on human-centric activities.Other datasets~\cite{talk2car, mono3dvg,talk2lidar} focus on 3DVG in outdoor autonomous driving scenes. However, there datasets tend to emphasize vehicles while neglect vulnerable road users (VRUs) related to the situational awareness of autonomous vehicles, such as various cyclists and pedestrians. Besides, sensors mounted on vehicle rooftops are prone to occlusion and limited field-of-view. In this paper, our work focuses on 3D visual grounding for outdoor monitoring scenarios where sensors are stationary relative to their surroundings.
\begin{figure*}[ht]
\centering
\includegraphics[width=1.0\textwidth]{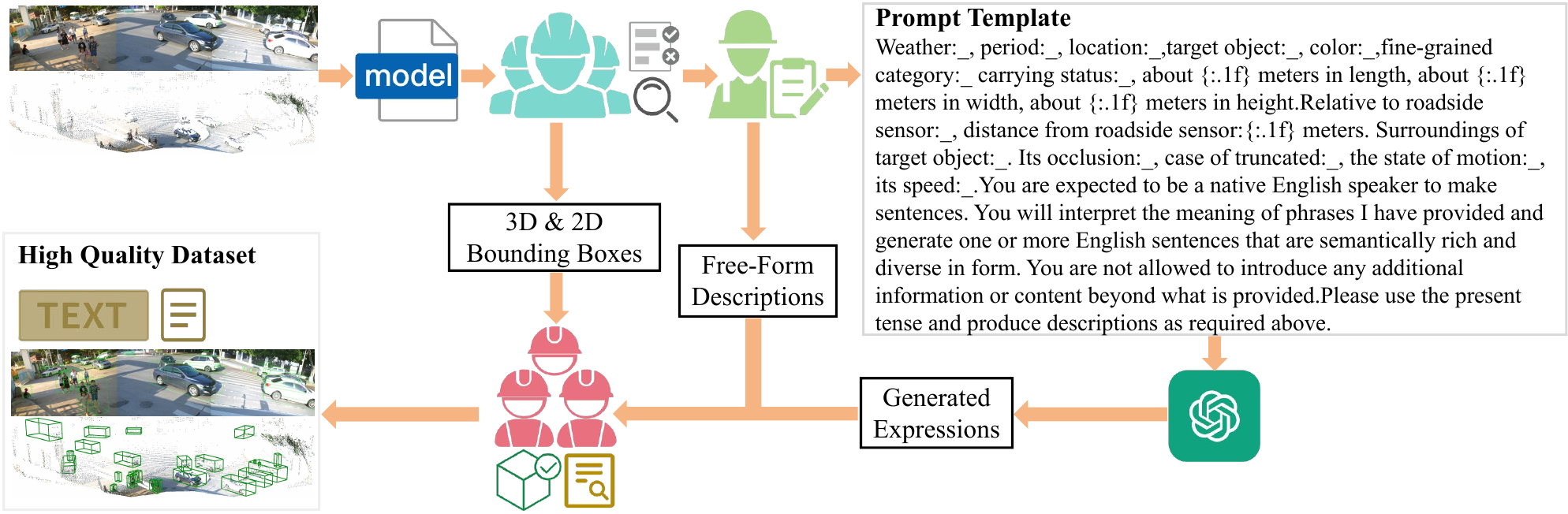}
\caption{\textbf{Our data collection pipeline.} For instance bounding boxes, we employ a pretrained object detector to generate preliminary annotations of objects, followed by manual inspection and refinement. For natural language expressions, annotators are required to provide diverse attributes for each object while labeling the 3D bounding boxes. Then, we obtain semantically rich and diverse textual descriptions by combining manual annotation with LLM-assisted generation.}
\label{fig:dataset_process}
\end{figure*}

\noindent\textbf{Infrastructure-based 3D Object Detection.}
3D object detection is fundamental to autonomous vehicles and intelligent transportation systems. Significant progress has been made in vehicle-centric 3D detection with the development of large-scale datasets~\cite{kitti, nuscenes, waymo} and advanced methods~\cite{pointrcnn, voxelnext, pointpillars,3dssd,virconv,voxelnet, bevfusion,detr3d}. Nevertheless, ego-vehicle sensors suffer from blind spots due to occlusions and limited perception range, which can compromise detection accuracy and driving safety. Roadside perception, offering far-reaching views and reduced blind spots, has therefore attracted increasing attention in intelligent transportation and smart city applications. Recent public datasets~\cite{rope3d,dair-v2x,r-livit,a9,rcooper} have facilitated the emergence of infrastructure-based 3D detection methods~\cite{infradet3d, bevheight, monogae}, such as CBR~\cite{cbr}, which enables calibration-free perception via decoupled feature reconstruction, MonoGAE~\cite{monogae}, which leverages implicit roadside ground information and high-dimensional semantic features for improved accuracy, and BEVHeight~\cite{bevheight}, which predicts object height and projects 2D features to 3D space. However, these approaches focus on detecting all candidate objects in a scene and do not support grounding a specific object based on textual descriptions. Motivated by this limitation, in this paper, we propose to explore the impact of natural language expressions with multiple attributes on infrastructure-based 3D object detection.

\section{MoniRefer Dataset}

As shown in Table~\ref{tab:1}, previous ScanRefer~\cite{scanrefer}, Sr3D~\cite{referit3d}, Nr3D~\cite{referit3d}, and SUNRefer~\cite{sunrefer}, concentrate on pre-scanned indoor scenes.Recent efforts such as STRefer~\cite{wildrefer}, LifeRefer~\cite{wildrefer}, Talk2Lidar~\cite{talk2lidar}, Talk2Car-3D~\cite{talk2lidar},NuPrompt~\cite{nuprompt}, and Mono3DRefer~\cite{mono3dvg} extend 3DVG to outdoor settings, but largely focus on autonomous driving or human-centric activities. To advance the boarder application of 3DVG in outdoor environments, we build a large-scale, multi-modal dataset captured by real-world infrastructure sensors. And we present detailed data acquisition in Sec.\ref{sec:3.1}, data annotations in Sec.\ref{sec:3.2}, and dataset statistics in Sec.\ref{sec:3.3}.

\begin{figure}[ht]
\centering
\includegraphics[width=0.46\textwidth]{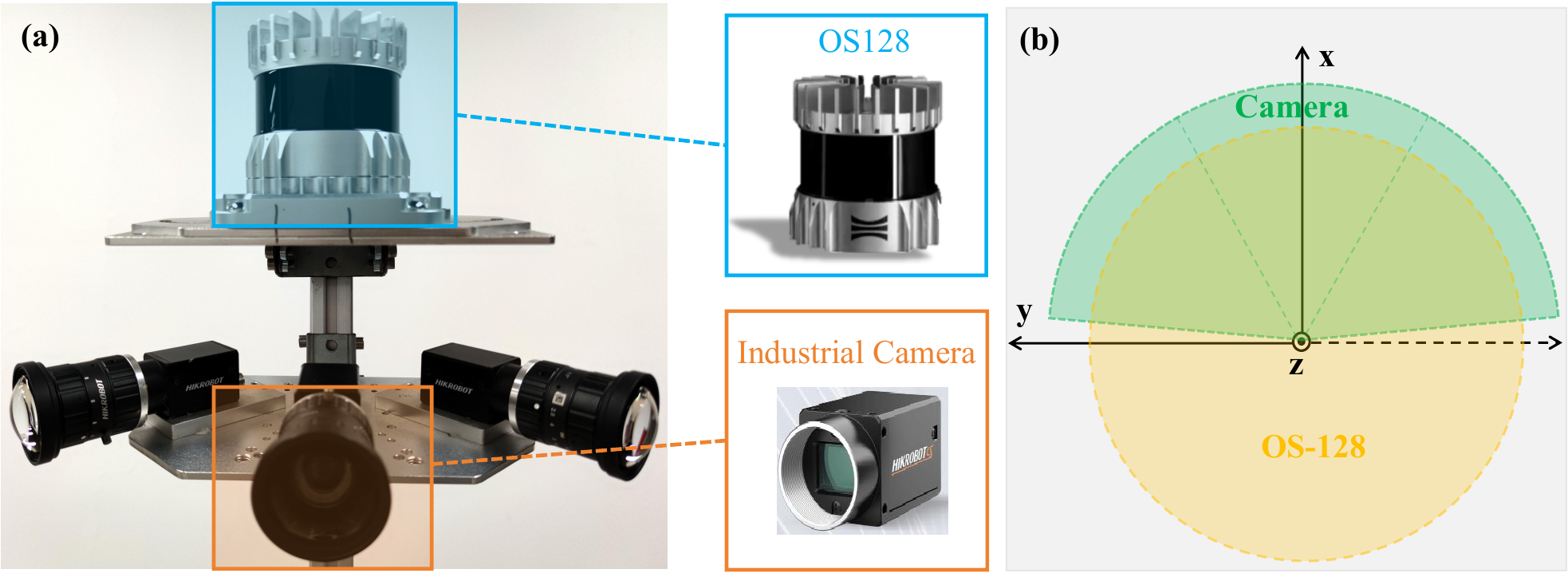}
\caption{\textbf{Sensors Setup and Annotation Area.}  (a) showcases the sensor layout of multi-source data acquisition platform. (b) demonstrates the effective annotation range composed of four sensors.}
\label{fig:data_collect}
\vspace{-4mm}
\end{figure}

\subsection{Data Acquisition}
\label{sec:3.1}
\paragraph{Sensors Setup.}
To capture large-scale multi-modal data across highly dynamic and complex traffic intersections, we deploy a multi-source sensing platform, as illustrated in Fig.~\ref{fig:data_collect}~(a). The system integrates four heterogeneous sensors: one 128-beam mechanical-scanning LiDAR and three identical industrial cameras. Detailed specifications are as follows:
\begin{itemize}
    \item \textbf{LiDAR}: OUSTER OS-1-128, 128 beams, 10Hz, $128 \times 1024$ resolution, $\text{360}^\circ$ horizontal FOV, $\text{-22.5}^\circ \sim \text{+22.5}^\circ$ vertical FOV, and $\leq 170\,$m range.
    \item \textbf{Industrial cameras}: HIKVISION MV-CA023-10GC with MVL-KF1228M-12MP lens, 10Hz, $1920 \times 1200$ resolution, $\text{-29.9}^\circ \sim \text{+29.9}^\circ$ horizontal FOV, $\text{-23.1}^\circ \sim \text{+23.1}^\circ$ vertical FOV, equipped with a $1/1.2''$ CMOS sensor.
\end{itemize}

\begin{figure*}[t]
\centering
\includegraphics[width=1.0\textwidth]{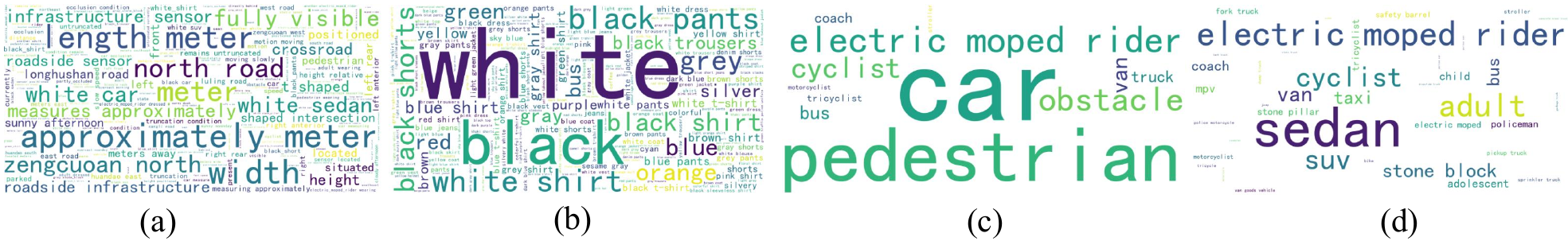}
\caption{(a)-(d) present word clouds of terms corresponding to unique words across the dataset, color, coarse-grained object categories, and fine-grained object categories in the MoniRefer dataset, respectively. The fonts size reflects the occurrence frequency of each term in the descriptions.}
\label{fig:statistic}
\vspace{-4mm}
\end{figure*}

\paragraph{Synchronization and Calibration.}
To achieve precise temporal alignment across heterogeneous sensors, we adopt the Precision Time Protocol (PTP)~\cite{2002ieee}, ensuring sub-millisecond synchronization accuracy.
High-quality multimodal data acquisition further requires accurate spatial calibration. Our pipeline involves three coordinate systems: the high-precision point cloud coordinate system, the camera coordinate system, and the LiDAR coordinate system. First, we estimate each camera's intrinsic parameters using OpenCV~\cite{opencv} with a set of chessboard-pattern images. Next, we obtain LiDAR extrinsics by jointly scanning the same scene with a high-precision 3D laser scanner and the LiDAR. We then compute camera extrinsics by detecting corresponding corners in both camera images and the high-precision point cloud using OpenCV. With all intrinsic and extrinsic parameters estimated, transformations among all sensor coordinate systems are fully established, enabling accurate cross-modal fusion.

\paragraph{Data Collection.}

To enhance the diversity and complexity of our dataset, we deployed our data collection platform at 15 distinct locations across 5 real-world traffic intersections, continuously collecting data over a period of one month. The resulting dataset ultimately comprises 12,085 frames, covering a wide variety of weather conditions (cloudy, overcast, sunny, windy), times of day (morning, afternoon, dusk), intersection types (crossroads, T-junctions, roundabouts), traffic densities (with or without traffic lights, crowded, normal, sparse), and geographic settings (urban, suburban, rural). To enable precise multi-modal fusion, we utilized the accurately calibrated intrinsic and extrinsic parameter matrices to transform the coordinate system and extract color information for each point in every LiDAR frame, yielding a high-fidelity pseudo-color point cloud representation. This extensive and richly annotated dataset not only supports infrastructure-level 2D/3D object detection for vehicles and vulnerable road users, but also surpasses existing 3D visual grounding datasets in both scale and environmental diversity.

\begin{figure*}[ht]
\centering
\includegraphics[width=1.0\textwidth]{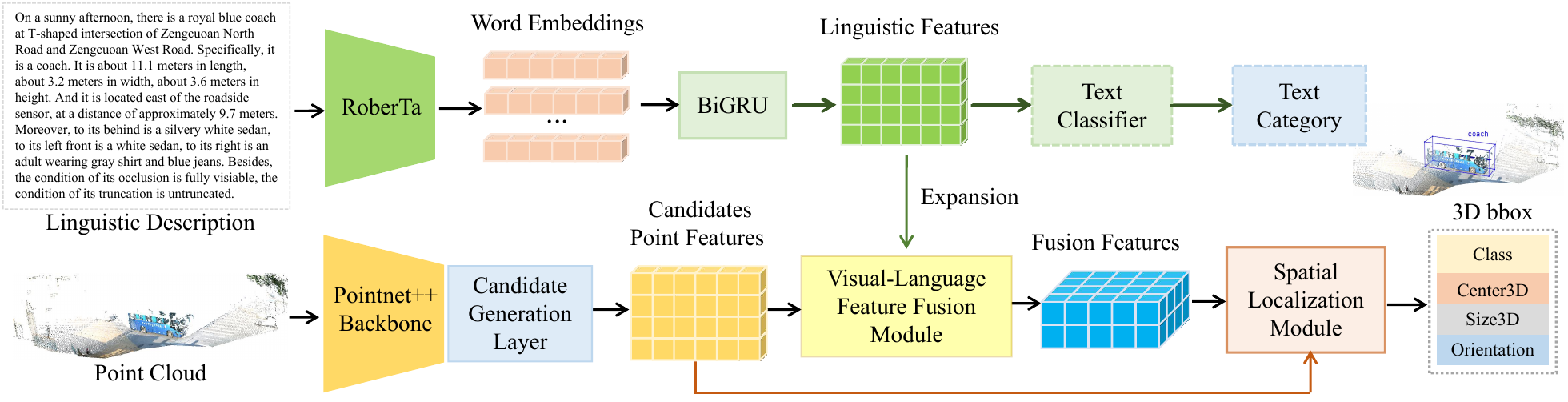}
\caption{\textbf{The overview of Moni3DVG framework.} We formulate the 3DVG task as identifying the candidate point that is closest to the target object center. Specifically, the visual and linguistic encoders extract candidate point representations and textual features from the point clouds and natural language description, respectively. These features are projected into a latent space, where effective cross-modal interaction and fusion are performed to yield enriched multi-modal representations. Finally, the spatial localization module leverages the fused multi-modal features to identify the candidate point nearest to the referred object center and output the 3D bounding box that best matches the referring expression.}
\label{fig:moni3DVG}
\vspace{-4mm}
\end{figure*}

\begin{figure*}[t]
\centering
\includegraphics[width=1.0\textwidth]{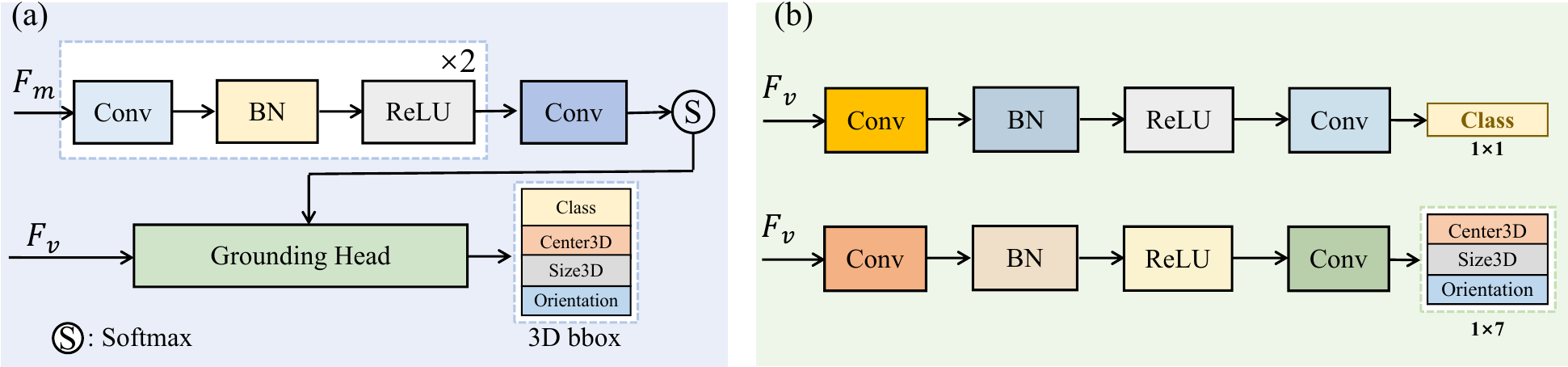}
\caption{Detail of spatial localization module~(a) and grounding head~(b).~$F_m$, and $F_v$ denotes multi-modal features and candidate point features, respectively.}
\label{fig:spatial_localization}
\vspace{-4mm}
\end{figure*}

\begin{figure}[t]
\centering
\includegraphics[width=0.46\textwidth]{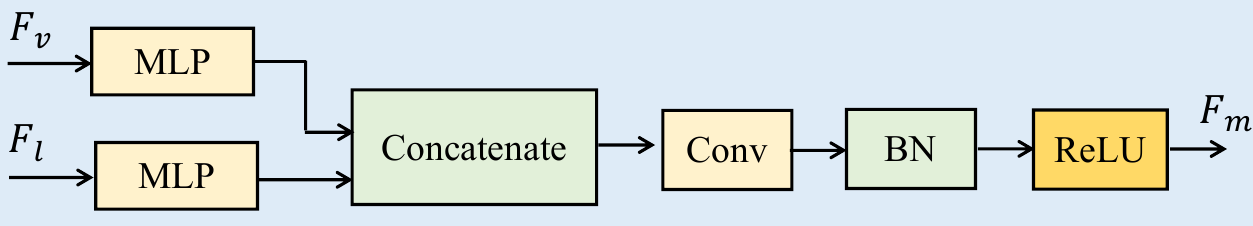}
\caption{Detail of visual-language feature fusion module. $F_v$, $F_l$ and $F_m$ represents visual, linguistic and fused multi-modal features, respectively. }
\label{fig:visual_lang_fusion}
\vspace{-4mm}
\end{figure}

\subsection{Data Annotations}
\label{sec:3.2}

\paragraph{Annotation Area.}
Leveraging precise time synchronization and multi-sensor calibration, data from LiDAR and RGB cameras can be jointly annotated, substantially improving labeling accuracy, particularly for distant objects. We define an effective annotation range of $\text{50m}$, which corresponds to the area simultaneously observable by LiDAR and cameras, as shown in Fig.~\ref{fig:data_collect}~(b). All objects of interest within this range are systematically annotated to ensure consistent coverage across the dataset.

\paragraph{Annotation Rule.}

Our dataset includes 12 object categories: car, van, bus, coach, truck, pedestrian, cyclist, tricycle, motorcyclist, electric moped rider, stroller and obstacle. Each object is labeled with a 7-DoF 3D bounding box $(x, y, z, l, w, h,\theta)$, where $(x, y, z)$ denotes the center coordinates, $(l, w, h)$ represents length, width, and height, and $\theta$ indicates orientation. Beyond geometric annotations, we provide more than 10 types of attributes information for each object, such as period, location, color, position, orientation, surrounding, motion, speed, and so on.

\paragraph{Annotation Process.}
To cover the dataset efficiently while reducing inter-frame redundancy, we sample frames from the raw data at intervals of one every ten frames. We adopt a semi-automatic annotation pipeline to get 3D boxes and natural language descriptions, as depicted in Fig.~\ref{fig:dataset_process}: ~\textbf{For 3D bounding boxes}, we first generate proposals using a pretrained object detection model, followed by manual inspection and refinement through multiple quality validation cycles conducted by 17 trained annotators. This ensures that each bounding box accurately represents the corresponding real-world object. ~\textbf{For 2D bounding boxes}, we project the manually verified 3D bounding boxes onto the images of three views using the calibrated high-precision parameter matrices.~\textbf{For referring expressions}, we employ a hybrid human-and-LLM approach. First, annotators are required to provide more than ten critical attributes for each object in the scene while labeling the 3D bounding boxes. These attributes offer more discriminative cues for ambiguous objects, enabling more accurate localization. Then, annotators generate natural language descriptions based on these key attributes, such that each description uniquely identifies the target object within the 3D scene. To further enhance linguistic diversity and richness, these attributes are fed into a prompt template and passed to a large language model (ChatGPT) for textual expressions generation. All natural language expressions are manually verified by nine team members to ensure accuracy, uniqueness, and consistency. The entire annotation process spans 13 months. Additionally, all potentially sensitive information, such as vehicle license plates and human faces, are blurred using professional labeling tools to ensure privacy compliance. We provide more details about labeling process in the appendix.

\subsection{Dataset Statistics}
\label{sec:3.3}
Table~\ref{tab:1} presents an overview of our dataset. We collect 12,085 outdoor monitoring scenes using roadside infrastructure sensors equipped with LiDAR and cameras. In total, annotated objects include vehicles, vulnerable road users (VRUs), and generic obstacles within the 50\,meters effective range, covering 12 coarse-grained categories and 49 fine-grained categories. Since more detailed descriptions can provide discriminative features that distinguish the target from other objects within the same category, we generate natural language descriptions based on more than ten attributes for each object through a combination of manual annotation and ChatGPT-assisted generation.Finally, MoniRefer contains 411{,}128 natural language expressions for 136{,}018 objects, with an average of 126.78 words per description and a vocabulary of 8,478 unique words. Fig.~\ref{fig:statistic} illustrates different word clouds of our dataset,such as color and object category. Additional detailed statistics and analyses are provided in the supplementary materials.


\section{Methodology}
In this section, we first formally define a novel grounding task and provide an overview of our proposed framework in Section ~\ref{sec:overview}. Section ~\ref{sec:VFE} and Section ~\ref{sec:LFE} then describe the encoding of visual and textual features from the input point cloud and the linguistic query, respectively. Section ~\ref{sec:VLFF} presents our multi-modal feature fusion module, which facilitates effective interaction and fusion between visual and linguistic information. Section ~\ref{sec:SL} introduces the spatial localization module for identifying the target object corresponding to the input description. Finally, the loss function is detailed in Section ~\ref{sec:loss}.

\subsection{Overview}\label{sec:overview}
We introduce the task of 3D visual grounding for outdoor monitoring scenarios (see Fig.~\ref{fig:3dvg_task}), where the inputs include a LiDAR point cloud collected from roadside sensors and a diverse linguistic expression referring to a specific object in the scene. Each point encodes geometric information as well as additional appearance and optical attributes. The goal is to predict the 3D bounding box of the object that corresponds to the given description in the physical world.

Inspired by 3DSSD~\cite{3dssd}, we formulate the task as identifying the candidate point closest to the target object's center among a set of candidate points. To this end, we propose Moni3DVG~(see Fig.~\ref{fig:moni3DVG}), an end-to-end framework composed of four core components: (1) a visual feature encoder;(2) a linguistic feature encoder; (3) a multi-modal feature fusion module;(4) a spatial localization module. The visual encoder processes the point cloud to generate representations for candidate points, while the linguistic encoder extracts semantic features from the textual description. These features are then projected into a shared embedding space, where effective cross-modal interaction and fusion are performed to produce enriched multi-modal representations. Finally, the spatial localization module leverages these fused features to predict the target object's 3D bounding box.

\subsection{Visual Feature Encoder}\label{sec:VFE}
We adopt an enhanced PointNet++ backbone~\cite{pointnet++} with the fusion sampling strategy and the candidate generation layer from 3DSSD~\cite{3dssd} to process the point cloud and extract global features for candidate points. The candidate generation layer outputs a set of center point features denoted as $F_v \in \mathbb{R}^{M \times C_v}$, where $M$ is the maximum number of candidate points and $C_v$ is the feature dimension for each point. These features serve as the visual representation of each candidate point for downstream multi-modal fusion.

\subsection{Linguistic Feature Encoder}\label{sec:LFE}
For the natural language descriptions, we employ a pretrained RoBERTa-base model~\cite{roberta} to generate word embeddings $F_w \in \mathbb{R}^{L \times C_t}$, where $L$ denotes the number of tokens in the referring expressions and $C_t$ is the vector dimension of each token. The word embeddings are sequentially processed by a bi-directional GRU~\cite{gru} to capture contextual dependencies within the sentence. The final hidden state of the bi-directional GRU serves as the aggregated linguistic features $F_l \in \mathbb{R}^{1 \times C_l}$ representing the entire textual query, which will be aligned with visual features in the subsequent fusion stage.

\subsection{Visual-Language Feature Fusion}\label{sec:VLFF}
To bridge the semantic gap between visual and textual features, visual-language feature fusion module is proposed. As shown in Fig.~\ref{fig:visual_lang_fusion}, we first project $F_v$ and $F_l$ into a shared latent space. Then each candidate point feature is concatenated with the corresponding textual feature and processed through a multi-layer perceptron (MLP) to generate fused multi-modal representations $F_{m} \in \mathbb{R}^{M \times C_m}$, where M indicates the number of candidate points and $C_m$ represents the dimensionality of the fused features. This fusion module enables effective cross-modal interaction, allowing the network to integrate geometric, appearance, optical and linguistic cues for accurate target localization.

\subsection{Spatial Localization}\label{sec:SL}
The spatial localization module aims to identify the candidate point closest to the center of the ground-truth bounding box corresponding to the textual query. The fused features $F_{m}$ are passed through a single-layer MLP to produce raw scores indicating the likelihood of each candidate point being the target center (Fig.~\ref{fig:spatial_localization} (a)). A softmax function normalizes these scores into localization confidence values $S \in [0,1]$ for all $M$ points. The point with the highest confidence is selected, and its feature is further processed by grounding head to produce the final 3D bounding box corresponding to the input description~(Fig.~\ref{fig:spatial_localization} (b)).

\subsection{Loss Function}\label{sec:loss}
The overall training process is supervised by five different losses:
a reference loss $\mathcal{L}_{ref}$ for referred object localization, a language classification loss $\mathcal{L}_{lang}$ for text to object category classification, and a classification loss $\mathcal{L}_{cls}$, a regression loss $\mathcal{L}_{reg}$ as well as shifting loss $\mathcal{L}_{shift}$ for 3D bounding box prediction.
Specifically, we adopt the classification loss $\mathcal{L}_{cls}$, regression loss $\mathcal{L}_{reg}$, and shifting loss $\mathcal{L}_{shift}$ from 3DSSD~\cite{3dssd} to supervise candidate points classification, bounding box regression, and center shift prediction, respectively. The language classification loss $\mathcal{L}_{lang}$ is a multi-class cross-entropy loss for predicting the target category based on given description. The reference loss $\mathcal{L}_{ref}$ is formulated as a cross-entropy loss for referred target identification. Finally, the total loss function is computed as a weighted sum of these losses:
\begin{equation}
    \mathcal{L} = \lambda_{1} \mathcal{L}_{cls} + \lambda_{2} \mathcal{L}_{reg} + \lambda_{3} \mathcal{L}_{shift} + \lambda_{4} \mathcal{L}_{lang} + \lambda_{5} \mathcal{L}_{ref},
\end{equation}
where $\lambda_i$ is a balancing factor for weighting the contribution of each loss term.





\section{Experiments}

\begin{table*}[ht]
\caption{\textbf{Comparison results on the MoniRefer validation set.}  We measure percentage of predictions whose IoU with the ground truth boxes are greater than 0.25 and 0.5. We report scores of Acc@0.25 and Acc@0.5 for the \textit{Unique} subset, the \textit{Multiple} subset, and \textit{Overall}. The baselines include two-stage and one-stage approaches across different sensor modalities. "Photog." denotes photogrammetry. ${*}$ means training with ground-truth boxes.}
\label{tab:2}
\centering
  \setlength{\tabcolsep}{4pt}
    \resizebox{1.\linewidth}{!}{
\begin{tabular}{lccc|cc|cc|cc}
\Xhline{2\arrayrulewidth}
\multicolumn{4}{c|}{}                                         & \multicolumn{2}{c|}{Unique}     & \multicolumn{2}{c|}{Multiple}    & \multicolumn{2}{c}{Overall}     \\
Method                 & venue       & Sensor    & Type      & Acc@0.25       & Acc@0.5       & Acc@0.25       & Acc@0.5        & Acc@0.25       & Acc@0.5        \\
\Xhline{2\arrayrulewidth}
CatRandGT              & -           & LiDAR+RGB & Two-stage & 100            & 100           & 19.26          & 19.23          & 26.66          & 26.63          \\
3DSSDBest              & -           & LiDAR+RGB & Two-stage & 91.36          & 68.8          & 79.48          & 63.62          & 80.56          & 64.09          \\
3DSSDRand              & -           & LiDAR+RGB & Two-stage & 13.93          & 11.88         & 5.42           & 4.92           & 6.2            & 5.56           \\
ScanRefer~\cite{scanrefer}              & ECCV2020    & RGB-D     & Two-stage & 7.25           & 2.43          & 5.48           & 3.41           & 5.64           & 3.32           \\
Instancerefer$^{*}$~\cite{instancerefer}      & ICCV2021    & RGB-D     & Two-stage & 53.45          & 11.06         & 45.01          & 7.79           & 45.78          & 8.09           \\
Cityrefer$^{*}$~\cite{cityrefer}          & NeurIPS2023 & UAV Photog.       & Two-stage & 53.54          & 11.08         & 44.31          & 8.99           & 45.16          & 9.18           \\
WildRefer~\cite{wildrefer}          & ECCV2024    & LiDAR+RGB & One-stage & 66.98          & 16.28         & 60.47          & 13.95          & 61.06          & 14.16          \\
\textbf{Moni3DVG(Our)} & -           & LiDAR+RGB & Two-stage & \textbf{76.72} & \textbf{57.9} & \textbf{60.43} & \textbf{51.47} & \textbf{61.92} & \textbf{52.06} \\
\Xhline{2\arrayrulewidth}
\end{tabular}
}

\end{table*}

\begin{table*}[ht]
\caption{\textbf{Performance comparison on the MoniRefer validation set across different distance ranges.} We report Acc@0.25 and Acc@0.5 for \textit{Near} (0–10m), \textit{Medium} (10–30m), and \textit{Far} (30–50m) subsets, as well as \textit{Overall}. "Photog." means photogrammetry. ${*}$ indicates training with ground-truth boxes.}
\label{tab:3}
\centering
  \setlength{\tabcolsep}{4pt}
    \resizebox{1.\linewidth}{!}{
\begin{tabular}{lcc|cc|cc|cc|cc}
\Xhline{2\arrayrulewidth}
\multicolumn{3}{c|}{}                                         & \multicolumn{2}{c|}{Near (0-10m)} & \multicolumn{2}{c|}{Medium (10-30m)} & \multicolumn{2}{c|}{Far (30-50m)} & \multicolumn{2}{c}{Overall}     \\
Method           & Sensor    & Type      & Acc@0.25       & Acc@0.5        & Acc@0.25         & Acc@0.5         & Acc@0.25       & Acc@0.5        & Acc@0.25       & Acc@0.5        \\
\Xhline{2\arrayrulewidth}
CatRandGT                         & LiDAR+RGB & Two-stage & 29.55          & 29.45          & 25.77            & 25.75           & 27.75          & 27.75          & 26.66          & 26.63          \\
3DSSDBest                         & LiDAR+RGB & Two-stage & 62.54          & 42.45          & 83.41            & 70.74           & 84.84          & 56.01          & 80.56          & 64.09          \\
3DSSDRand                        & LiDAR+RGB & Two-stage & 4.58           & 4.06           & 7.11             & 6.50            & 3.9            & 3.03           & 6.2            & 5.56           \\
ScanRefer~\cite{scanrefer}                  & RGB-D     & Two-stage & 14.52          & 8.66           & 4.92             & 2.93            & 0.73           & 0.2            & 5.64           & 3.32           \\
Instancerefer$^{*}$~\cite{instancerefer}          & RGB-D     & Two-stage & 58.62          & 18.15          & 50.9             & 7.74            & 13.51          & 0.6            & 45.78          & 8.09           \\
Cityrefer$^{*}$~\cite{cityrefer}          & UAV Photog.      & Two-stage & 61.72          & 24.52          & 49.49            & 7.97            & 12.79          & 0.58           & 45.16          & 9.18           \\
WildRefer~\cite{wildrefer}            & LiDAR+RGB & One-stage & 64.18          & 20.41          & 70.64            & 12.17           & 26.83          & 7.03           & 61.06          & 14.16          \\
\textbf{Moni3DVG(Our)}         & LiDAR+RGB & Two-stage & \textbf{52.78} & \textbf{36.87} & \textbf{65.16}   & \textbf{57.88}  & \textbf{56.73} & \textbf{41.69} & \textbf{61.92} & \textbf{52.06} \\
\Xhline{2\arrayrulewidth}
\end{tabular}
}

\end{table*}

\subsection{Experimental Setup}

\paragraph{Implementation Details.}
We split our dataset into train/val/test sets with 289,634, 66,981 and 54,513 samples respectively.
Our experiments are conducted on six NVIDIA RTX 3090 GPUs using the open-source OpenPCDet~\cite{openpcdet} codebase. The network is trained for 60 epochs with a batch size of 10, using the Adam optimizer with a learning rate of $1\text{e-4}$ and weight decay of $1\text{e-4}$. The learning rate is reduced by a factor of 10 after 35 and 45 epochs. The weights of loss terms, $\lambda_{1}$, $\lambda_{2}$,$\lambda_{3}$, $\lambda_{4}$, and $\lambda_{5}$ are empirically set to 10, 10, 10, 1, and 1, respectively.

\paragraph{Evaluation Metrics.}
Following prior 3DVG works~\cite{scanrefer, instancerefer, butd_detr} in indoor scenes, we adopt Acc@K ($K = 0.25$ or $0.5$) as the evaluation metric. Acc@K measures the percentage of predicted bounding boxes whose Intersection over Union (IoU) with the ground-truth exceeds the threshold $K$.

\paragraph{Baselines.}
As our work is the first to explore 3D visual grounding in outdoor surveillance scenarios using multi-modal data from infrastructure sensors, there is no related work for absolutely fair comparison. To explore the difficulty of this task and verify the effectiveness of our approach, we therefore design several baselines and validate these methods under a unified metrics. 

\textbf{\textit{CatRandGT}}. This baseline randomly selects a ground-truth box of the same category as the referred object, serving to estimate the inherent difficulty of the task and dataset. It evaluates grounding performance when only object category information is available, while all other contextual and linguistic cues are disregarded.

\textbf{\textit{3DSSDRand}}. This baseline randomly chooses a predicted bounding box with the correct semantic class from proposals generated by 3DSSD detector~\cite{3dssd}. In contrast to CatRandGT, its performance is highly dependent on the quality of proposals from object detectors.

\textbf{\textit{3DSSDBest}}. This baseline selects the predicted proposal with the highest IoU with the ground-truth box among all 3DSSD predictions, providing the upper bound on how well the two-stage methods work for our novel task.

\textbf{\textit{Existing 3DVG Methods}}. In addition to the above baselines, We also evaluate several state-of-the-art 3DVG methods based on RGB-D~\cite{scanrefer, instancerefer}, UAV Photogrammetry~\cite{cityrefer}, and LiDAR~\cite{wildrefer} data, as shown in Table~\ref{tab:2}. 

To investigate the impact of additional information conveyed by input description beyond object category cues, we evaluate performance of these baselines on "unique" and "multiple" subsets in Table~\ref{tab:2}. The "unique" subset includes samples in which there is only one object of the relevant category matches expression, whereas the "multiple" subset contains multiple ambiguous instances with the same category. Furthermore, to examine task difficulty and the effect of sensor perception range, we introduce a distance-based analysis with three levels: Near (0–10m), Medium (10–30m), and Far (30–50m), as reported in Table~\ref{tab:3}. The "Near", "Medium", and "Far" subsets corresponds to cases where the distance between object center and infrastructure sensor falls within 0-10m, 10-30m, and 30-50m, respectively.

\begin{table*}[ht]
\caption{\textbf{Ablation study of Moni3DVG with different input modalities on the validation set of the MoniRefer dataset.} We report the percentage of predictions whose IoU with the ground-truth bounding boxes exceeds 0.25 and 0.5. The \textit{Unique} subset contains samples with no distracting objects, while the \textit{Multiple} subset refers to cases with multiple objects of the same category.}

\label{tab:4}
\centering
  \setlength{\tabcolsep}{8pt}
    \resizebox{1.\linewidth}{!}{
\begin{tabular}{cc|cc|cc|cc}
\Xhline{2\arrayrulewidth}
                  &                            & \multicolumn{2}{c|}{Unique}     & \multicolumn{2}{c|}{Multiple}    & \multicolumn{2}{c}{Overall}     \\
Method            & Data                       & Acc@0.25       & Acc@0.5       & Acc@0.25       & Acc@0.5        & Acc@0.25       & Acc@0.5        \\
\Xhline{2\arrayrulewidth}
Moni3DVG          & xyz                        & 76.46          & 58.4          & 58.65          & 49.97          & 60.28          & 50.75          \\
Moni3DVG          & xyz+intensity              & 75.95          & 54.25         & 59.12          & 51.36          & 60.66          & 51.62          \\
Moni3DVG          & xyz+rgb                    & 77.55          & 60.35         & 59.31          & 50.66          & 60.98          & 51.55          \\
\textbf{Moni3DVG} & \textbf{xyz+rgb+intensity} & \textbf{76.72} & \textbf{57.9} & \textbf{60.43} & \textbf{51.47} & \textbf{61.92} & \textbf{52.06} \\
\Xhline{2\arrayrulewidth}
\end{tabular}}

\end{table*}

\begin{table*}[ht]
\caption{\textbf{Ablation study of Moni3DVG with different input modalities on the validation set of the MoniRefer dataset under varying distance ranges.} Based on the distance between the referred object and the sensor, samples are divided into \textit{Near} (0–10m), \textit{Medium} (10–30m), \textit{Far} (30–50m) subsets. Acc@0.25 and Acc@0.5 are reported for each subset as well as \textit{Overall}.}
\label{tab:5}
\centering
\setlength{\tabcolsep}{8pt}
\resizebox{1.\linewidth}{!}{
\begin{tabular}{cc|cc|cc|cc|cc}
\Xhline{2\arrayrulewidth}
                  &                            & \multicolumn{2}{c|}{Near (0-10m)} & \multicolumn{2}{c|}{Medium (10-30m)} & \multicolumn{2}{c|}{Far (30-50m)} & \multicolumn{2}{c}{Overall}                        \\
Method            & Data                       & Acc@0.25       & Acc@0.5        & Acc@0.25         & Acc@0.5         & Acc@0.25       & Acc@0.5     &
Acc@0.25       & Acc@0.5                           \\
\Xhline{2\arrayrulewidth}
Moni3DVG          & xyz                        & 53.15          & 40.2           & 61.85            & 55.2            & 60.2           & 41.85       & 60.28          & 50.75                             \\
Moni3DVG          & xyz+intensity              & 50.89          & 38.3           & 63.57            & 56.49           & 57.41          & 43.46       &
60.66          & 51.62                             \\
Moni3DVG          & xyz+rgb                    & 51.1           & 37.53          & 64               & 57.31           & 57.37          & 40.37       & 60.98          & 51.55                             \\
\textbf{Moni3DVG} & \textbf{xyz+rgb+intensity} & \textbf{52.78} & \textbf{36.87} & \textbf{65.16}   & \textbf{57.88}  & \textbf{56.73} & \textbf{41.69}  & \textbf{61.92} & \textbf{52.06}                    \\
\Xhline{2\arrayrulewidth}
\end{tabular}
}

\end{table*}

\subsection{Quantitative Analysis}
As shown in Table~\ref{tab:2}, we evaluate the performance of our model and several baselines on the val set of MoniRefer. CatRandGT achieves 100\% accuracy on the \textit{Unique} subset but drops sharply to 19\% on the \textit{Multiple} subset, highlighting the challenge posed by multi-object ambiguity in our task. Although 3DSSDRand performs poorly overall, it achieves higher accuracy in \textit{Unique} than in \textit{Multiple} subset, indicating that category information alone can be sufficient when there is only a single object of its class in the scene. A notable performance gap remains between our end-to-end Moni3DVG and 3DSSDBest, suggesting substantial room for improvement the match between language expression and the visual concepts. To further demonstrate the novelty of our task and the effectiveness of our approach, we also evaluate several state-of-the-art 3DVG approaches based on different sensor on MoniRefer. ScanRefer, originally designed for indoor 3DVG, performs poorly due to its limited sensing range and the mismatch in data distribution. CityRefer, derived from InstanceRefer and using ground-truth boxes,  exhibits similar performance across all subsets. WildRefer, which targets human-centric indoor and outdoor scenarios, is constrained by its accuracy threshold. In contrast, Moni3DVG, specifically designed for outdoor monitoring scenarios, demonstrates strong adaptability and effectiveness under complex and high dynamic traffic environments.

As reported in Table~\ref{tab:3}, CatRandGT performs slightly better on the \textit{Near} subset than on the \textit{Medium} and \textit{Far} subsets, while 3DSSDRand performs best on the \textit{Medium} subset, indicating that model is biased toward medium-range objects. Both ScanRefer and InstanceRefer, which rely on RGB-D data collected in indoor environments, show decreasing performance as distance increases and perform best on the \textit{Near}(0–10 m) subset. Although CityRefer employs ground-truth bounding boxes, it struggles to handle roadside monitoring scenarios due to inconsistent data distributions. WildRefer, which leverages data from RGB cameras and LiDAR with a maximum sensing range of 30 m, presents a sharp performance drop in the \textit{Far} (30–50 m) subset, and its localization accuracy is further limited by the evaluation threshold. In contrast, our method outperforms all baselines beyond 3DSSDBest, demonstrating superior performance. Notably, it not only models contextual relations among objects to disambiguate same category instances but also effectively associates semantic information between textual descriptions and scene context, resulting in more accurate target localization. Additional results and analyses on the test set are provided in the supplementary material.

\subsection{Qualitative Analysis}
\begin{figure*}[ht]
\centering
\includegraphics[width=1.0\textwidth]{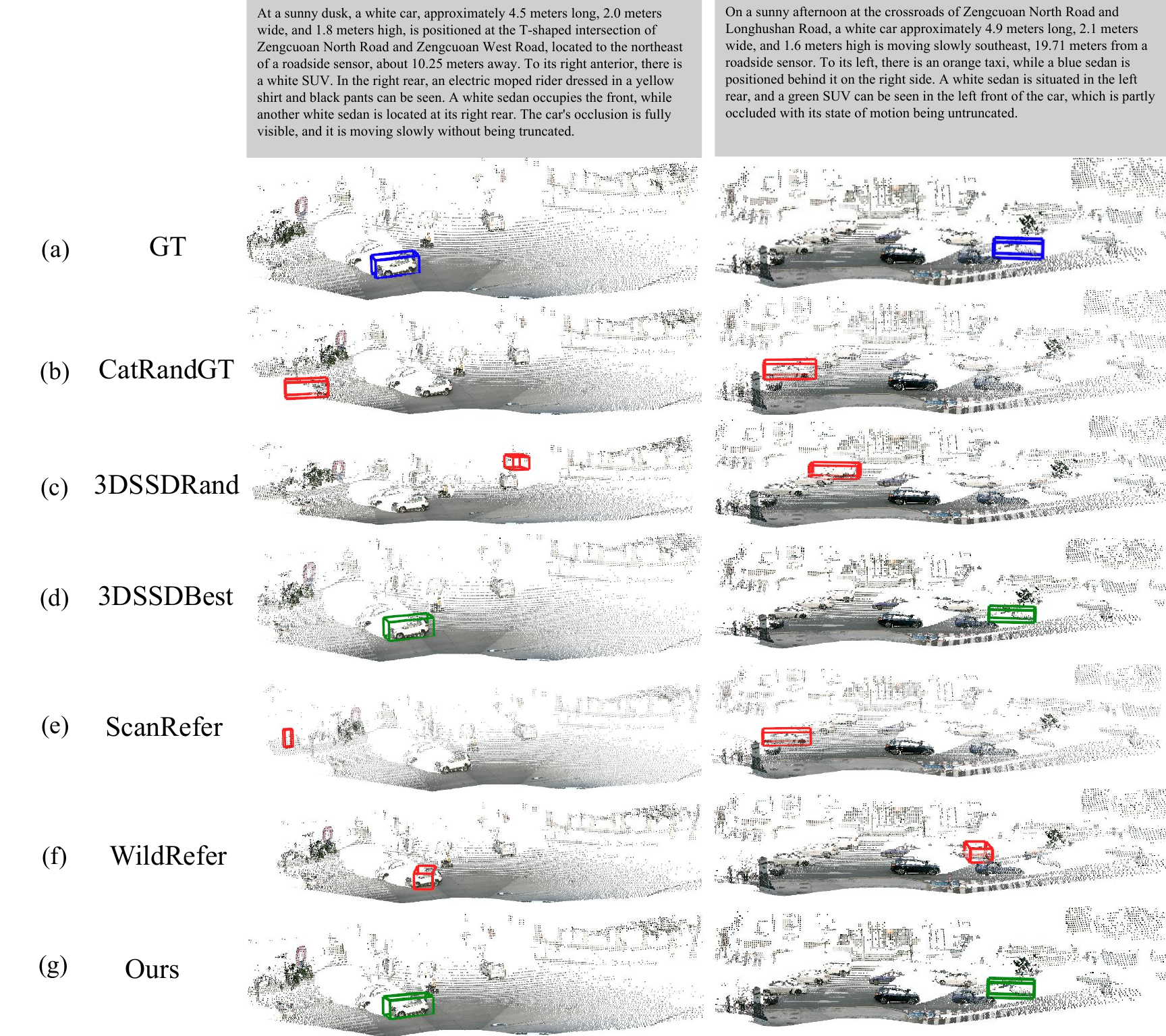}
\caption{\textbf{Qualitative results from baseline methods and our Moni3DVG on MoniRefer dataset.} Blue, green and red boxes denote the ground-truth, right prediction, and wrong prediction, respectively.}
\label{fig:comparasion}
\end{figure*}

Fig.~\ref{fig:comparasion} presents visualization results of CatRandGT, 3DSSDRand, 3DSSDBest, ScanRefer, WildRefer, and our Moni3DVG on samples randomly selected from two different scenarios. In these scenes containing multiple objects of the same category, CatRandGT and 3DSSDRand are unable to identify the referred object when relying solely on object category information. ScanRefer fails to predict targets beyond the RGB-D camera’s sensing range (5 m), as the objects lie within the \textit{Medium}(10–30 m) subset . Although WildRefer leverages both LiDAR and RGB inputs, it is designed for human-centric activity scenarios and struggles to accurately localize objects beyond the human category. In contrast, both 3DSSDBest and our method produce correct predictions, where 3DSSDBest serves as the upper bound of our two-stage framework. Notably, our method not only demonstrates a strong capability in handling spatial relations to disambiguate same category objects, bur also effectively models semantic correlations between scene context and textual descriptions through multi-modal fusion and interaction, resulting in more accurate target localization. More detailed qualitative analyses are provided in the appendix.

\subsection{Ablation Studies}
We conduct an ablation study to analyze the contribution of different point cloud features to our model performance. Table~\ref{tab:4} reports Acc@0.25 and Acc@0.5 on the \textit{Unique}, \textit{Multiple}, and \textit{Overall}, while Table~\ref{tab:5} presents the results on the \textit{Near}, \textit{Medium}, \textit{Far}, and \textit{Overall}. As shown in Table~\ref{tab:4}, the first row corresponds to using only geometric features(xyz), and introducing optical features (intensity) in the second row leads to consistent performance improvements on the \textit{Overall}. Incorporating appearance features(rgb) in the third row further enhances the results. When all feature modalities(xyz+rgb+intensity) are combined, the model achieves the best performance, demonstrating the complementary nature of geometric, optical, and appearance features. A similar trend is observed in Table~\ref{tab:5}, where adding either appearance (rgb) or optical (intensity) information improves localization accuracy over the geometry-only baseline, and the best results can be improved by approximately 1\%-2\% when all features(xyz+rgb+intensity) are used. Results in both tables all demonstrate that integrating multiple feature information consistently enhances the overall target localization performance of the model. More additional ablation analyses are provided in the supplementary material.

\section{Conclusion}
In this paper, we explore a novel 3D visual grounding for outdoor monitoring scenarios, which aims to localize target objects in 3D space based on natural language expressions and multi-modal visual data captured by roadside infrastructure sensors. We further contribute a high quality large-scale multi-modal dataset for roadside-level 3D visual grounding, which contains 411,128 descriptions for 136,018 objects. Moreover, we propose Moni3DVG, an end-to-end framework that integrates geometric, appearance, and optical information from point clouds and images, enabling effective multi-modal feature learning and accurate 3D object localization. Extensive experiments demonstrate the effectiveness and state-of-the-art performance of our approach. Finally, we hope our dataset and method will advance 3D vision-language research and foster the development of intelligent infrastructure systems and embodied agents capable of human-like recognition. 

\clearpage
{
    \small
    \bibliographystyle{ieeenat_fullname}
    \bibliography{monirefer}
}

  
\end{document}